\documentclass[runningheads]{llncs}
\usepackage{marvosym}
\usepackage[T1]{fontenc}
\usepackage{graphicx}
\usepackage{caption} 
\usepackage{xcolor}
\usepackage{ulem} 
\usepackage{appendix}
\usepackage{hyperref}
\usepackage{graphicx}
\usepackage{subfigure}
\usepackage{multirow}
\usepackage{tabularx}
\usepackage{booktabs}
\usepackage{adjustbox}
\usepackage{xcolor}
\usepackage{ulem} 
\usepackage{appendix}
\usepackage{hyperref}
\usepackage{graphicx}
\usepackage{subfigure}
\usepackage{amssymb}
\usepackage{multirow}
\usepackage{amsmath}
\usepackage{amssymb}
\usepackage{tabularx}
\usepackage{booktabs}
\usepackage{adjustbox}
\usepackage{caption} 
\usepackage{amsfonts} 
\hypersetup{
    colorlinks=true,
    linkcolor=blue,
    citecolor=blue,
    urlcolor=blue
}
\definecolor{race}{HTML}{D46C76}
\definecolor{jeff}{HTML}{2196F3}
\definecolor{Sully}{HTML}{E09BA2}
\definecolor{Labradorintro1}{HTML}{D46C76}
\definecolor{Sullyintro1}{HTML}{95DA69}
\definecolor{Labrador}{HTML}{339EF5}
\definecolor{BushJunior}{HTML}{339EF5}
\definecolor{BushSenior}{HTML}{E09BA2}
\definecolor{BarackObama}{HTML}{339EF5}
\definecolor{jg}{HTML}{9FC2F5}
\definecolor{mylegal}{RGB}{241, 172, 106}
\usepackage{tikz}

\begin{document}
%
\title{MUSE: Integrating Multi-Knowledge for Knowledge Graph Completion}
\titlerunning{MUSE: Integrating Multi-Knowledge for Knowledge Graph Completion}
%
\author{Pengjie Liu\inst{1}\textsuperscript{(\Letter)}\orcidID{0000-0001-8044-0134} 
}
\institute{School of Computer Science and Engineering, Southern University of Science and Technology, Shenzhen, China\\ 
\email{12031109@mail.sustech.edu.cn}\\
}
\maketitle              
\begin{abstract}
Knowledge Graph Completion (KGC) aims to predict the missing [\textit{{relation}}] part of (head entity)\textbf{{\(\xrightarrow{[relation]}\)}}(tail entity) triplet. 
Most existing KGC methods focus on single features (e.g., relation types) or sub-graph aggregation. However, they do not fully explore the Knowledge Graph (KG) features and neglect the guidance of external semantic knowledge.
To address these shortcomings, we propose a knowledge-aware reasoning model (MUSE), which designs a novel multi-knowledge representation learning mechanism for missing relation prediction. Our model develops a tailored embedding space through three parallel components: 1) \textbf{Prior Knowledge Learning} for enhancing the triplets' semantic representation by fine-tuning BERT; 2) \textbf{Context Message Passing} for enhancing the context messages of KG; 3) \textbf{Relational Path Aggregation} for enhancing the path representation from the head entity to the tail entity. The experimental results show that MUSE significantly outperforms other baselines on four public datasets, achieving over \textbf{5.50\% H@1} improvement and \textbf{4.20 \% MRR} improvement on the NELL995 dataset. The code and datasets will be released via https://github.com/SUSTech-TP/ADMA2024-MUSE.git.
\end{abstract}
\section{Introduction}
\label{sec:intro}
\begin{figure}{
\centering 
\subfigure[\label{fig:1a}
\textbf{Limited Information Set (LIS) Scenario}. When we predict the relation between the \textbf{\color{Labradorintro1}Labrador} and \textbf{\color{Sullyintro1}Sully}, the description knowledge guides MUSE to identify Sully is a dog. This recognition indicates that the correct relation is \textbf{Breed} rather than Food. 
 ]
 {
            \includegraphics[width=1\textwidth]{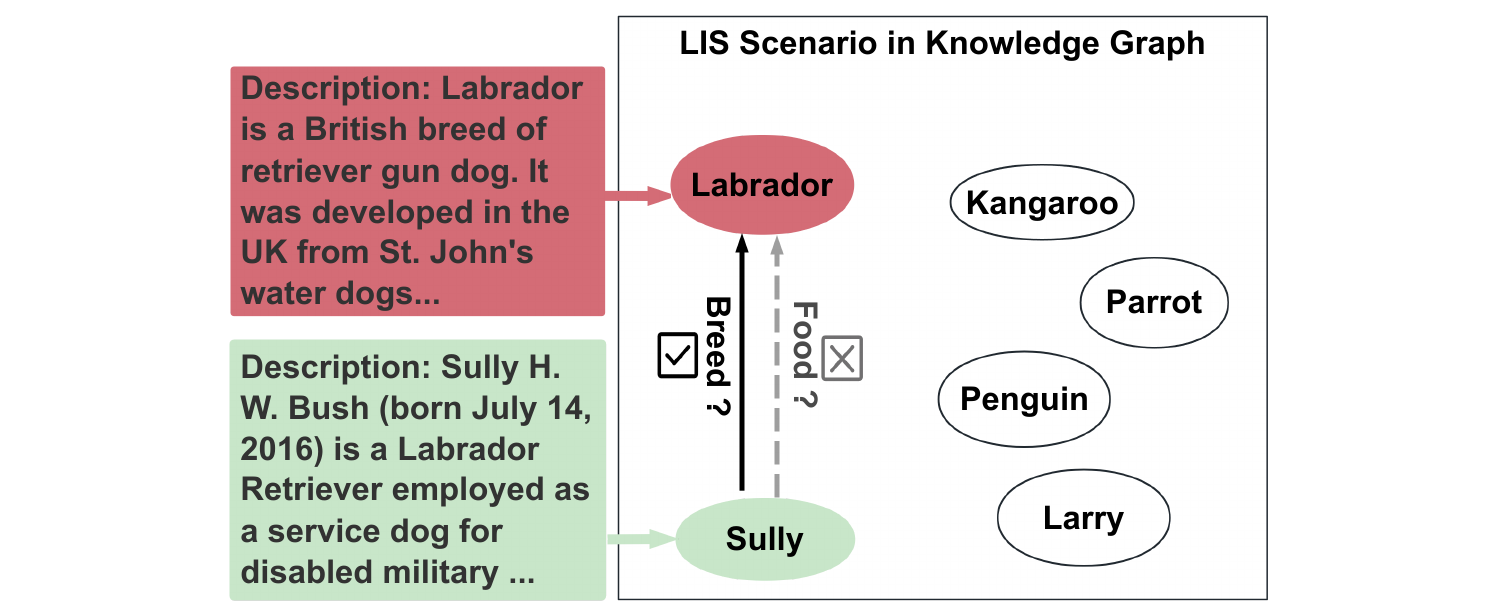
            }
            }
    \hfill
  \subfigure[\label{fig:1b}
  \centering 
 \textbf{Rich Information Set (RIS) Scenario}. When we predict Bush Senior is the \textbf{Father} of \textbf{\color{BushJunior}Bush Junior} or \textbf{\color{Sully}Barack Obama}, 
 their descriptions highlight many similarities in presidential terms and political careers. However, they take different paths to the Bush Senior. Bush Junior's paths contain \{(Bush Senior)\textbf{\color{BushJunior}{\(\xrightarrow{}\)}}(Barbara Pierce Bush)\textbf{\color{BushJunior}{\(\xrightarrow{}\)}}(Bush Junior)\} and 
 \{(Bush Senior)\textbf{{\(\xrightarrow{}\)}}(USA
Government)\textbf{{\(\xrightarrow{}\)}}(Bush Junior)\}, while Barack Obama's path is 
\{(Bush Senior)\textbf{{\(\xrightarrow{\color{Sully}}\)}}(USA
Government)\textbf{{\(\xrightarrow{\color{Sully}}\)}}(Barack Obama)\} only. Leveraging the multi-knowledge reasoning mechanism, MUSE realizes that Bush Senior is the father of Bush Junior not Barack Obama.
  ]{
           \includegraphics[width=1\textwidth] {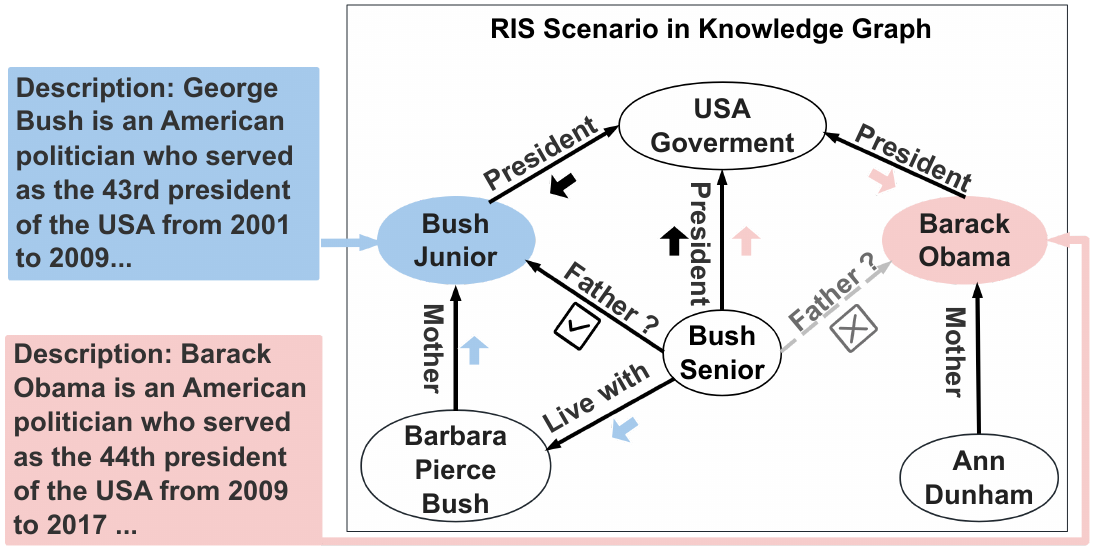}
            }
    \caption{\label{fig:1}
    Two Example Cases of Relation Prediction in the LIS and RIS Scenarios. 
    }
    }
\end{figure}
Knowledge Graph (KG) encapsulate triplet data in a structured format, containing head and tail entities along with their relations~\cite{bordes2011learning,fusion2,ali}. 
Nevertheless, most real-world KGs suffer from incomplete datasets.
Therefore, extensive researchers have proposed various Knowledge Graph Completion (KGC) models to predict missing relations over entity representation learning~\cite{kgc_review_1,kgc_review_2,nayyeri2021trans4e}.

Existing KGC methods have two main categories: single-knowledge-based models~\cite{transe,neurip,Random}, and multi-knowledge-fusion-based models~\cite{zhang2020relational,kgbert,pathcon,doloers}. The single-knowledge-based KGC models, such as TransE~\cite{transe}, TransH~\cite{transh}, TransD~\cite{transd}, and TransR~\cite{transr}, typically rely on individual features within the KG.
These models leverage the embeddings of head and tail entities to compute relation scores for potential candidates. By employing various translation functions, they determine the relation scores and select the candidate with the highest score for relation prediction.
Many KG paths involve more than two entities from the head entity to the tail entity. Consequently, research has focused on path ranking algorithms \cite{Random} and rule mining methods \cite{neurip,drum} to improve the search efficiency for these multi-entity paths.
Besides, inspired by the GNNs in sub-graph representation learning, some KGC methods adopt the node-based message passing mechanism to propagate and aggregate the features among connecting nodes~\cite{conv,inductive,semi}. 

Different from the above methods that learn some specific single features, recent multi-knowledge-fusion-based KGC models explore the fusion of textual description and the graph structure~\cite{BERT-2019,fusion1,fusion2,text2}, or the combination of connected nodes and paths~\cite{pathcon,distmult}. 
However, These two KGC models both suffer the long-tail problem in the entity and relation prediction task, especially when dealing with sparsely distributed graph nodes. The long-tail issue significantly increases the difficulty of KGC tasks and reduces prediction accuracy~\cite{longtail_v1,Towards}. 

In this paper, we propose a knowledge-aware reasoning model, MUSE, which can train a tailored embedding space for the missing relation prediction. MUSE conducts a multi-knowledge reasoning mechanism through \textbf{Prior Knowledge Learning}, \textbf{Context Message Passing}, and \textbf{Relational Path Aggregation}. 
Specifically, during the prior knowledge learning, we 
fine-tune BERT through a relation classification task. Then we employ this fine-tuned BERT checkpoint to encode the description of given entity pairs and 
initial the graph to explore the sub-graph topology. Besides, MUSE aggregates the neighbor entity representation through the context message passing.
Meanwhile, our model enhances the path representation by reasoning and concating the entities, and relations on each path.
As illustrated in Figure~\ref{fig:1a}, we inject the prior knowledge into the entity representation when the topology information is limited.
For the RIS scenario in Figure~\ref{fig:1b}, the entity descriptions are highly similar and difficult to predict the answer. 
Therefore, we pay more attention to the context messages and path knowledge to reason the correct relation.
The experimental results on the NELL995 dataset demonstrate that MUSE outperforms other KGC models by more than \textbf{5.50\% H@1} in the relation prediction task. 
Additionally, our model has achieved \textbf{1.00 H@3} on both the WN18 and WN18RR datasets, highlighting its strong predictive capabilities.
Further analysis reveals that MUSE provides an effective multi-knowledge reasoning mechanism that can effectively and accurately enhance the entity semantic representation.
\vspace{-2mm}
\section{Related Work}
Knowledge Graph Completion (KGC) methods have two paradigms: single-knowledge-based models, and multi-knowledge-fusion-based models~\cite{zhang2020relational}. Besides, with the development of the Pre-trained Language Models (PLMs) in representation learning, many researchers attempt to enhance the Knowledge Graph (KG) representation by injecting the external semantic knowledge~\cite{kgbert}.
\subsection{Single-Knowledge-Based KGC Models}
The single-knowledge-based KGC models include the embedding-based, path-based, and Neural Networks-based (GNNs-based) methods, as they only leverage one kind of KG feature for prediction. Specifically, some embedding-based techniques are based on translation function, which treats entities as points in a continuous space and each relation as the translation function in the space. Their goals are to make the translated head entity be close to the tail entity in the same space~\cite{transe,transr,transh,transd}. 
Another embedding-based strategy includes multi-linear and bi-linear models, which compute semantic similarity through matrix dot products performed in real or complex spaces~\cite{complex}.

Then the path-based paradigm leverages connecting paths between head/tail entities for the unknown information prediction. Specifically, the random walk reasoning algorithm learns each enumerated relational path as the one-hot vector through a random walk algorithm, which is predicted by a trainable classifier~\cite{Random}. Rule mining methods, such as NeuralLP~\cite{neurip} and DRUM~\cite{drum} mainly learn probabilistic logic rules to weight different paths to achieve more accurate predictions. Besides, reasoning over Graph Neural Networks (GNNs) has developed as another potential framework for predicting the unknown paths of knowledge graphs~\cite{pathcon}.
\subsection{Multi-Knowledge-Fusion-Based KGC Models}
Previous representation learning research has mainly focused on isolated single knowledge within KGs. However, recent multi-knowledge-fusion-based methods capture more features from the sub-graph topology and the entity descriptions.
PathCon~\cite{pathcon} aggregates knowledge by combining context messages and paths within sub-graph topologies. 
The sub-graph topological information can be learned by the node-based message passing, or the relational message passing mechanism. Node-based message passing methods update the node's embedding using aggregation functions, involving information exchange with neighbor nodes.
Text-representation-based methods aim to semantically encode textual knowledge~\cite{BERT-2019,text2,pre_1,cherniavskii-etal-2022-acceptability}.
Existing approaches typically commence with the fine-tuning of pre-trained models. Entities and relations are then initialized. Subsequently, the KG representation is updated via corresponding learning functions.
\subsection{Pre-training Language Models}
The PLMs have drawn more researchers' attention because they are effective ways to enhance the KG's representation~\cite{kgbert,liu2024semdrsemanticawaredualencoder}.
There are two main PLM techniques: feature-based, and fine-tuning-based methods~\cite{kgbert}.
Some feature-based work applies the Word2Vec, Glove, and ELMo algorithms to embed the graph~\cite{node2vec}. 
Recent researchers then use the pre-trained model
architecture and parameters to learn the contextual embedding as initialization of translation-based KG embedding models~\cite{doloers}. The COMET~\cite{COMET} employs GPT to create tail phrase tokens in a common sense knowledge base, utilizing given head phrases and types of relations. ERNIE~\cite{ERNIE} leverages the integration of information-rich entities from knowledge graphs to improve the effectiveness of language representation in BERT.
Differently, KG-BERT~\cite{kgbert} inputs names or descriptions of entities and relations to fine-tune BERT to calculate the corresponding plausibility scores of triplets based on the translation function.

\section{Methodology}
\label{sec:Methodology}
In this section, we present the framework of MUSE during relation prediction.
\subsection{Preliminary of Knowledge Graph Completion}
\begin{figure*}[t]
\centerline{\includegraphics[width=0.9\textwidth]{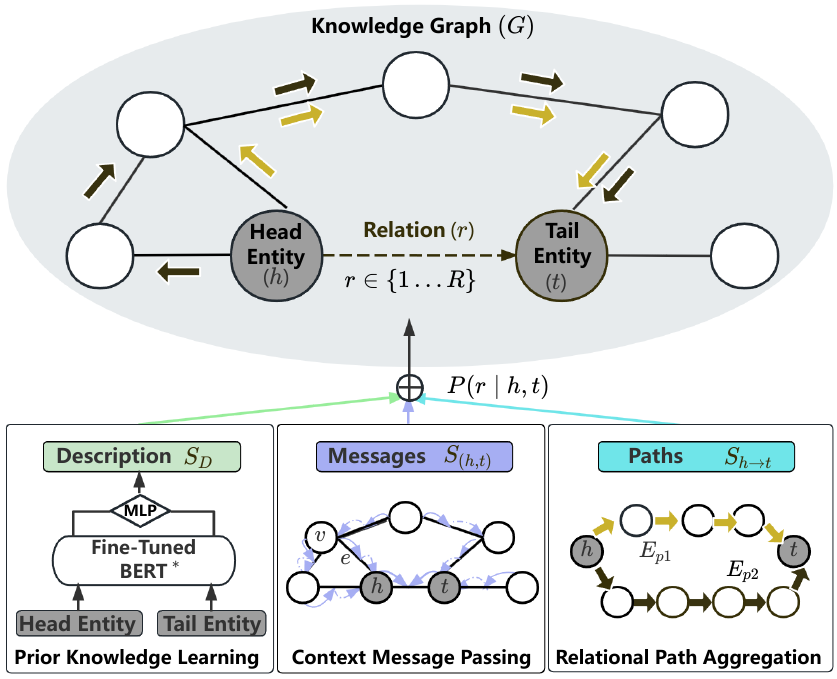}}
\caption{\label{1}The Architecture of MUSE Framework in Knowledge Graph Completion.}
\end{figure*}
\begin{figure}[t]
\centering
\includegraphics[width=0.6\textwidth]{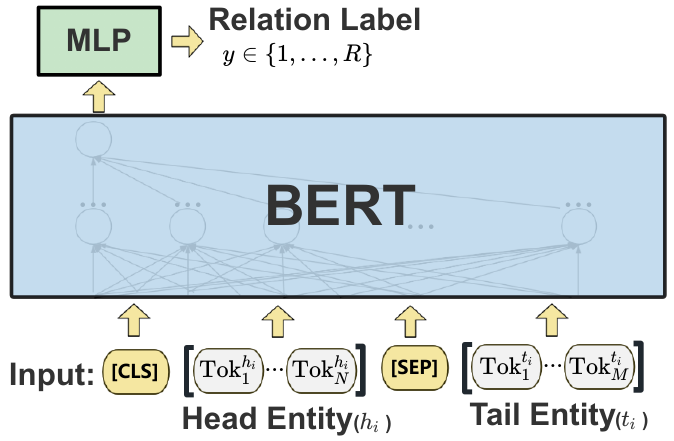} 
\caption{\label{fig:pkp}Illustration of the Prior Knowledge Learning. We fine-tune BERT on the datasets: FB15k-237, WN18, WN18RR, and NELL995, respectively.}
\end{figure}
The Knowledge Graph Completion (KGC) aims to predict the relation (${r}$) based on the head entity (\textbf{${h}$}), tail entity (\textbf{${t}$}) within a given triplet (\textbf{${\tau}$}). 
In the knowledge graph (\textbf{${G}$}), we also consider the entity as node (${v}$), and the relation as edge (${e}$).

As shown in Figure~\ref{1}, MUSE consists of three modules: Prior Knowledge Learning, Context Message Passing, and Relational Path Aggregation. We first learn entity description representation~($S_{D}$) by employing the fine-tuned BERT~\cite{BERT-2019} model~(Sec. \ref{Sec.3.2}). Then MUSE establishes a graph attention network to enhance the context message representation~($S_{(h, t)}$) by reasoning neighbor nodes~(Sec. \ref{Sec.3.3}). Besides, our model also captures the relational knowledge~($S_{h \rightarrow t}$) by aggregating all the connected paths from the head entity to tail entity~(Sec. \ref{Sec.3.4}).

Finally, MUSE is trained to predict
the relation via leveraging the integrated multi-knowledge using the following loss: 
\begin{equation}
\small{\mathcal{L_{\tau}}=\sum_{(h, r, t) \in \mathcal{T}} \operatorname{CrossEntropy}(r, P(r \mid h, t)),}
\end{equation}
where ${P(r \mid h, t)}$ denotes the probability to predict correct relation and it is calculated by aligning three knowledge representation learning as:
\begin{equation}
\small{P(r \mid h, t) = \operatorname{SoftMax}\left(S_{D} + S_{(h,t)} + S_{h \rightarrow t}\right).}
\end{equation}
\subsection{Prior Knowledge Learning\label{Sec.3.2}}
As illustrated in Figure~\ref{fig:pkp}, we fine-tune the BERT model through a classification task to enhance its capability in representing semantic knowledge.
For the ${i_{th}}$ triplet data $({\tau i})$, we tokenises the description of head entity (${h_{i}}$) and tail entity (${t_{i}}$) as ${D_{hi}=\left\{\operatorname{Tok}_{1}^{h_{i}}, \ldots, \operatorname{Tok}_{N}^{h_{i}}\right\}}$ and ${D_{t i}=\left\{\operatorname{Tok}_{1}^{t_{i}}, \ldots, \operatorname{Tok}_{M}^{t_{i}}\right\}}$. MUSE then employs the BERT model to encode this description set. It takes the [CLS] token as the final hidden state (${{C_{i}}}$) to calculate the triplet score (${S_{\tau i}}$):
\begin{equation}\label{eq:ft}
\small{{C}_{i}=\operatorname{BERT}([\text{CLS}], D_{hi}, [\text{SEP}], D_{ti})_{[CLS]},}
\end{equation}
\begin{equation}
\small{{S_{\tau i}= \mathrm{SoftMax} ({C_{i}}{W^{T}})},}
\end{equation}
where $W$ represents the learnable weights in the classification layer. We apply the triplet score and relation labels to calculate the fine-tuning loss (${{\mathcal{L}_{ft}}}$):
\begin{equation}\label{eq:input}
\small{{\mathcal{L}_{ft}=-\sum_{\tau \in \mathbb{G}} \sum_{j=1}^{R} y_{\tau}^{j} \log \left(S_{\tau i}\right)},}
\end{equation}
where ${y_{\tau}^{j}}$ denotes the relation indicator and ${j}$ represents any relation from candidates $\{1, \ldots, R\}$ within ${G}$. Specifically, for ${j = r}$, we have ${y_{\tau}^{j} = 1}$, otherwise if the ${j \neq r}$, we define ${y_{\tau}^{j} = 0}$. Then we leverage the fine-tuned BERT$^{*}$ to encode the entity description as:
\begin{equation}\label{eq:ft}
\small{{S}_{D}=\operatorname{MLP}[{\operatorname{BERT}^{*}(h)} \cdot {\operatorname{BERT}^{*}(t)}],}
\end{equation}
where MLP and ${[\cdot]}$ denote the multi-layer perceptron and concat function.
\subsection{Context Message Passing\label{Sec.3.3}}
MUSE follows the Pathcon~\cite{pathcon} model and designs an edge-based message passing mechanism to further enhance the sub-graph representation. 
During $d$ iteration time, given the context representation of edge (${{S_{e}^d}}$), we can obtain the node's message representation (${{m_{v}^d}}$) as: 
\begin{equation}\label{eq:sum}
\small{{m_{v}^d=\sum_{e \in \mathcal{N}(v)} s_{e}^d},}
\end{equation}
where ${e \in \mathcal{N}(v)}$ represents the set of connected edges. In the next iteration ($d+1$), the context representation of edges $(s_e^{d+1})$ could be aggregated by the relation message passing: 
\begin{equation}
\small{{S_{e}^{d+1}=\sigma\left(\left[m_{e}^d \cdot m_{v}^d \cdot S_{e}^d\right] \cdot W_e^{d}+b_e^{d}\right)},}
\end{equation}
where ${\sigma(\cdot) }$, ${W_e^{d}}$, and $b_e^{d}$ denote the ${Relu}$ activation function, learnable transformation matrix, and bias of aggregation. Consequently, after $K$ iterations of message passing, the context message representation ${S_{(h, t)}}$ for the entity pairs $(h,t)$ can be calculated as:
\begin{equation}\label{eq:Sei}
\small{{S_{(h, t)}=\sigma\left(\left[m_{h}^{K} \cdot  m_{t}^{K}\right] \cdot W_e^{K}+b_e^{K}\right)}.}
\end{equation}

The message representation $(m_{h/t})$ of head or tail entity could be aggregated by the edge-based attention mechanism as:
\begin{equation}
{m_{h/t}^{K}=\sum_{e \in \mathcal{N}(v)} \alpha_{e} s_{e}^{K}},
\end{equation}
\begin{equation}\label{eq:attention}
\small{{\alpha_{e}=\frac{\exp \left(s_{e}^{T}\cdot  {\operatorname{BERT}^{*}}{(h/t)}\right)}{\sum_{e \in \mathcal{N}(v)} \exp \left(s_{e}^{T} \cdot  {\operatorname{BERT}^{*}}{(h/t)}\right)},}}
\end{equation}
where ${e \in \mathcal{N}(v)}$ represents the set of connected nodes. We employ the fine-tuned BERT${^{*}}$ checkpoint in Equation~\ref{eq:ft} to initialize their representation. 
\subsection{Relational Path Aggregation\label{Sec.3.4}}
Our model also emphasizes the importance of reasoning the relational paths among some similar entities. We first use the one-hot encoder to initial each path's representation (${E_{P}}$). Then we apply the context knowledge representation ${S_{(h,t)}}$ in Equation~\ref{eq:Sei} to calculate the attention score (${\alpha_{P}}$) as:
\begin{equation}
\small{{{\alpha}_{P}=\frac{\exp \left(E_{P}^{\top} \cdot S_{(h,t)}\right)}{\sum_{P \in \mathcal{P}_{h \rightarrow t}} \exp \left(E_{P}^{\top} \cdot S_{e}\right)}},}
\end{equation}
where the path set (${\mathcal{P}_{h \rightarrow t}}$) contains all the paths from the head entity to the tail entity. Then we can update the path knowledge representation ($S_{h \rightarrow t}$) as:
\begin{equation}
\small{{S_{h \rightarrow t}=\sum_{P \in \mathcal{P}_{h \rightarrow t}} \alpha_{P} E_{P}}.}
\end{equation}

\section{Experimental Methodology}
\label{sec:Experiment}
\begin{table*}[t]
\centering
 \captionsetup{size=small,skip=5pt}
\caption{The Statistics Details of Raw Dataset and Data Splitting in Our Experiments.}
\resizebox{0.75\linewidth}{!}{
\small
\begin{tabular}{lrrrr} 
\hline
                                         & \textbf{FB15k-237}   & \textbf{WN18}        & \textbf{WN18RR}      & \textbf{NELL995}      \\ 
\hline
\multicolumn{1}{l}{\textbf{Raw Dataset}} & \multicolumn{1}{l}{} & \multicolumn{1}{l}{} & \multicolumn{1}{l}{} & \multicolumn{1}{l}{}  \\ 
\hline
Relation Type                          & 237                  & 18                   & 11                   & 198                   \\
 Entity Type                              & 14,541               & 40,943               & 40,943               & 63,917                \\
 Entity Degree Expectation                & 37.4                   & 6.9                    & 4.2                    & 4.3                     \\
 Entity Degree Variance                   & 12,336.0               & 236.4                  & 64.3                   & 750.6                   \\ 
\hline
\multicolumn{1}{l}{\textbf{Data Splits}} & \multicolumn{1}{l}{} & \multicolumn{1}{l}{} & \multicolumn{1}{l}{} & \multicolumn{1}{l}{}  \\ 
\hline
Train Triplets                       & 272,115              & 141,442              & 86,835               & 137,465               \\
Valid Triplets                  & 17,535               & 5,000                & 3,034                & 5,000                 \\
Test Triplets                         & 20,466               & 5,000                & 3,134                & 5,000                 \\ 
\hline
\multicolumn{1}{l}{\textbf{Testing Scenarios}}   & \multicolumn{1}{l}{} & \multicolumn{1}{l}{} & \multicolumn{1}{l}{} & \multicolumn{1}{l}{}  \\ 
\hline
LIS Scenario~(\%)~            & 2                    & 7                    & 21                   & 31                    \\
RIS Scenario~(\%)~               & 98                   & 93                   & 79                   & 69                    \\
\hline
\end{tabular}
}
\label{tab:statistics}
\end{table*}
In this section, we outline the experimental settings of MUSE and other baselines.
\subsection{Datasets}
We conduct evaluations of MUSE on four public datasets widely used in Knowledge Graph Completion (KGC) task: FB15k-237~\cite{observed}, WN18~\cite{bordes2011learning}, WN18RR~\cite{Dettmers_Minervini_Stenetorp_Riedel_2018}, and NELL995~\cite{nell}. More details of dataset statistics are list
in Table~\ref{tab:statistics}. 
We note substantial differences in the entity degree expectations and variances across four datasets. NELL995 has a mean degree of 4.2 and a variance of 750.6, whereas FB15k-237 has much higher mean degree of 37.4 and a variance of 12,336.

\textit{Testing Scenarios}.
We have followed KICGPT~\cite{KICGPT} and established the Limited Information Set (\textbf{LIS}) scenario and Rich Information Set (\textbf{RIS}) scenario according to the entity degree. Specifically, LIS entity degree ranges from 0 to 3, whereas RIS entity degrees of 3 or more.
Besides, the degree of an entity is calculated as the maximum of either the sum of its in-degree and out-degree, or the number of paths from this entity to other tail entities ($\text{entity degree} = \mathrm{max}{\{\text{(in-degree + out-degree), paths}}\}$). Higher LIS percentage indicates more long-tail entities in the sparser distribution. The NELL995 dataset possesses the highest percentage of long-tail entities among the four datasets, reaching 31\%.
\subsection{Baselines}
We evaluate MUSE and several KGC baselines on the relation prediction task.

\textit{Single-Knowledge-Based models}: TransE~\cite{transe}, ComplEx~\cite{complex}, DistMult~\cite{graphvite}, RotatE~\cite{rotate}, and QuatE~\cite{quate} are embedding-based methods. Their main difference is the type of continuous space for entities. DRUM~\cite{drum} is a path representation learning method capturing path features using probabilistic logical rules.

\textit{Multi-Knowledge-Fusion-Based models}: PathCon~\cite{pathcon} is one of the latest SOTA KGC methods, which can learn both the context information and relational path features from the head entity to the target tail entity. KG-BERT~\cite{kgbert} is a method that enhances textual features of entities by fine-tuning the BERT.

\textit{Evaluation Metrics}.
The official KGC evaluation metrics include Mean Reciprocal Rank (MRR), HIT@1 (H@1), and HIT@3 (H@3). The H@1 is our main evaluation, and each MUSE experiment is repeated 3 times and we report the results as the average performance with the corresponding standard deviation.
\subsection{Implementation Details}
For TransE, ComplEx, DistMult, RotatE, QuatE, and DRUM, we set the embedding dimensions at 400 and training epoch is 1000. And we follow the hyper-parameter setting of the KG-BERT\footnote{\url{ https://github.com/yao8839836/kg-bert}} and PathCon\footnote{\url{https://github.com/hwwang55/PathCon}} in our experiments.

For the MUSE implementation, we start from the Bert-base-uncased\footnote{\url{https://huggingface.co/google-bert/bert-base-uncased}} and fine-tune it in prior knowledge learning.
Specifically, we limit each entity description to 512 tokens and fine-tune for 10 epochs.
MUSE then follows the optimal parameter setting used by PathCon to extract contextual and path features.
In our experiments, the datasets <FB15k-237, WN18, WN18RR, NELL995> are configured as follows: context layers <2, 3, 3, 2>, max path length <3, 3, 4, 5>, learning rate <1e-4, 1e-4, 5e-4, 1e-4>. 
We employ the Adam optimizer with a batch size of 128 and conduct 60 training epochs, setting the hidden layer dimensions to 64. During context message passing, we set the number of iterations to $K=2$. All experiments are conducted on 2 NVIDIA RTX 3090ti GPUs.

\section{Evaluation Results}
\begin{table*}[!t]
\centering
\caption{Relation Prediction in the General Scenarios. The best results are highlighted in \textbf{bold}, and the best results of the baseline are \uline{underlined}. }
\resizebox{1\linewidth}{!}{
\small
\begin{tabular}{l|ccc|ccc|ccc|ccc} 
\hline
\multicolumn{1}{c|}{\multirow{2}{*}{Methods}} & \multicolumn{3}{c|}{FB15k-237}                                                                    & \multicolumn{3}{c|}{WN18}                                                                & \multicolumn{3}{c|}{WN18RR}                                                              & \multicolumn{3}{c}{NELL995}                                                              \\ 
\cline{2-13}
\multicolumn{1}{c|}{}                         & MRR                                  & H@1                         & H@3                          & MRR                         & H@1                         & H@3                          & MRR                         & H@1                         & H@3                          & MRR                         & H@1                         & H@3                          \\ 
\hline
TransE                                        & 0.966                                & 0.946                       & 0.984                        & 0.971                       & 0.955                       & 0.984                        & 0.784                       & 0.669                       & 0.870                        & 0.841                       & 0.781                       & 0.889                        \\
ComplEx                                       & 0.924                                & 0.879                       & 0.970                        & 0.985                       & 0.979                       & 0.991                        & 0.840                       & 0.777                       & 0.880                        & 0.703                       & 0.625                       & 0.765                        \\
DistMult                                      & 0.871                                & 0.802                       & 0.933                        & 0.786                       & 0.584                       & 0.987                        & 0.847                       & 0.779                       & 0.891                        & 0.634                       & 0.524                       & 0.714                        \\
RotatE                                        & 0.970                                & 0.951                       & 0.980                        & 0.984                       & 0.979\uline{}               & 0.986\uline{}                & 0.799                       & 0.735                       & 0.823                        & 0.729                       & 0.691                       & 0.756                        \\
QuatE                                         & 0.974\uline{}                        & 0.958\uline{}               & 0.988\uline{}                & 0.981                       & 0.975                       & 0.983                        & 0.823                       & 0.767                       & 0.852                        & 0.752\uline{}               & 0.706\uline{}               & 0.783\uline{}                \\
DRUM                                          & 0.959                                & 0.905                       & 0.958                        & 0.969                       & 0.956                       & 0.980                        & 0.854\uline{}               & 0.778\uline{}               & 0.912\uline{}                & 0.715                       & 0.640                       & 0.740                        \\
PathCon                                       & \uline{0.979}                        & \uline{0.964}               & \uline{0.994}                & \uline{0.993}               & \uline{0.988}               & \uline{0.998}                & \uline{0.974}               & \uline{0.954}               & 0.994                        & 0.896                       & \uline{0.844}               & 0.941                        \\
KG-BERT                                       & 0.973                                & 0.953                       & 0.993                        & 0.992                       & 0.987                       & 0.997                        & \textbf{0.991}              & \textbf{0.983}              & \uline{0.999}                & \uline{0.897}               & 0.821                       & \uline{0.970}                \\ 
\hline
\multirow{2}{*}{\textbf{MUSE}}                & \textbf{0.985}                       & \textbf{0.974}              & \textbf{0.997}               & \textbf{0.995}              & \textbf{0.992}              & \textbf{1.000}               & 0.986                       & 0.975                       & \textbf{1.000}               & \textbf{0.939}              & \textbf{0.899}              & \textbf{0.981}               \\
                                              & \multicolumn{1}{l}{\textbf{± }0.000} & \multicolumn{1}{l}{± 0.001} & \multicolumn{1}{l|}{± 0.000} & \multicolumn{1}{l}{± 0.001} & \multicolumn{1}{l}{± 0.001} & \multicolumn{1}{l|}{± 0.000} & \multicolumn{1}{l}{± 0.001} & \multicolumn{1}{l}{± 0.002} & \multicolumn{1}{l|}{± 0.000} & \multicolumn{1}{l}{± 0.002} & \multicolumn{1}{l}{± 0.003} & \multicolumn{1}{l}{± 0.002}  \\
\hline
\end{tabular}
}
\label{tab:overall}
\end{table*}
In this section, the experimental results are compared and analyzed to verify the validity and improvement of MUSE.
\subsection{Overall Performance}
This subsection evaluates the results of MUSE and baselines in the relation prediction task. We find that our model has almost achieved the best performance across all datasets, including the FB15K-237, WN18, WN18RR, and NELL995.

As shown in Table \ref{tab:overall}, MUSE improves H@1 by 1\%, 0.4\%, and 5.5\% over baseline models on the FB15K-237, WN18, and NELL995 datasets, respectively.
Our model has already achieved 1.0 H@3 accuracy on both WN18 and WN18RR datasets.
On the NELL995 dataset, our model achieves a consistent improvement in H@3 by 4.0\% over PathCon and 1.1\% over KG-BERT, along with a 4.2\% gain in MRR.
Besides, MUSE performs better in the Knowledge Graph (KG) with more sparsely distributed nodes. Specifically, our model achieves the most significant increase compared to PathCon on the WN18RR and NELL995 datasets.
They contain the highest Limited Information Set (LIS) ratios, exceeding 21\% and 31\%. Such consistent growth supports that MUSE has strong reasoning capability to address the long-tail problem.
\begin{table*}[t]
\small
\centering
\caption{Effectiveness of Ablation Models in the Relation Prediction Task.}
\resizebox{0.9\linewidth}{!}{
\begin{tabular}{l|l|c|c|c|c} 
\hline
\multirow{2}{*}{Knowledge } & \multirow{2}{*}{Methods}        & FB15k-237      & WN18           & WN18RR                                                                    & NELL995         \\ 
\cline{3-6}
                          &                                 & H@1            & H@1            & H@1                                                                       & H@1             \\ 
\hline
- 
                          & Backbone Model     & 0.943          & 0.951          & 0.661                                                                     & 0.779           \\ 
\hline
\multirow{3}{*}{Single}   & w/ Prior Knowledge     & 0.929          & 0.795          & 0.835                                                                     & 0.828           \\
                          & w/ Relational Path  & 0.957          & 0.971          & 0.897                                                                     & 0.685           \\
                          & w/ Context Message      & 0.961          & 0.927          & 0.894                                                                     & 0.815           \\ 
\hline
\multirow{3}{*}{Dual}     & w/ Prior Knowledge \& Relational Path     & 0.973\uline{}  & 0.988\uline{}  & \textcolor[rgb]{0.2,0.2,0.2}{0.96}\textcolor[rgb]{0.2,0.2,0.2}{5}\uline{} & 0.892\uline{}   \\
                          & w/ Prior Knowledge \& Context Message \space & 0.965          & 0.954          & \textcolor[rgb]{0.2,0.2,0.2}{0.90}\textcolor[rgb]{0.2,0.2,0.2}{3}         & 0.874           \\
                          & w/ Relational Path \& Context Message   & 0.964          & 0.988\uline{}  & 0.954                                                                     & 0.844           \\ 
\hline
\textbf{All}              & \textbf{MUSE}                   & \textbf{0.974} & \textbf{0.992} & \textbf{0.975}                                                            & \textbf{0.899}  \\
\hline
\end{tabular}
 }
\label{tab:as}
\end{table*}

\subsection{Ablation Study}
MUSE has three parallel components: Prior Knowledge Learning, Context Message Passing, and Relational Path Aggregation. 
In this subsection, we conduct the ablation study to investigate the role played by each representation learning module. All results are illustrated in Table~\ref{tab:as}.

Generally, these three single knowledge indeed help to improve the effectiveness of our model in the relation prediction task. Such a phenomenon reveals that the representation learning of prior knowledge, relational path, and context message all can boost prediction performance. In datasets with a high LIS proportion like NELL995, the semantic features in prior knowledge effectively direct the model toward the correct relation. Conversely, in more RIS scenarios, MUSE primarily acquires knowledge from the graph's topological structure.

In addition, our experimental results show that dual-knowledge fusion models consistently outperform their single-knowledge counterparts. Specifically, the Prior Knowledge \& Relational Path model achieves 0.973 H@1 on the FB15k-237 dataset, compared to 0.929 for Prior Knowledge and 0.957 for Relational Path. MUSE surpasses all single and dual knowledge-based models with its multi-knowledge reasoning mechanism. This supports our research that integrating semantic knowledge into the graph enriches entity representation and significantly enhances relation prediction accuracy.
\begin{figure*}[!t]
    \centering 
    \subfigure[The Effectiveness of Fine-Tuning BERT in the Prior Knowledge.] { 
    \label{fig:pkla} 
     \includegraphics[width=0.42\linewidth]{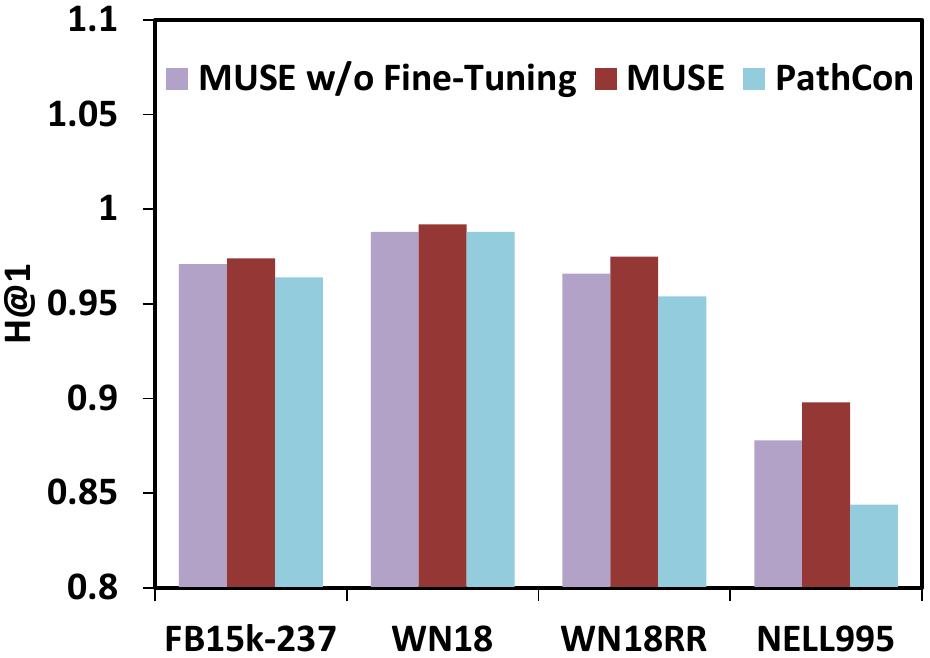}}
     \subfigure[The Effectiveness of Attention Mechanism in the Context
Message.] { 
     \label{fig:pklb}
    \includegraphics[width=0.42\linewidth]{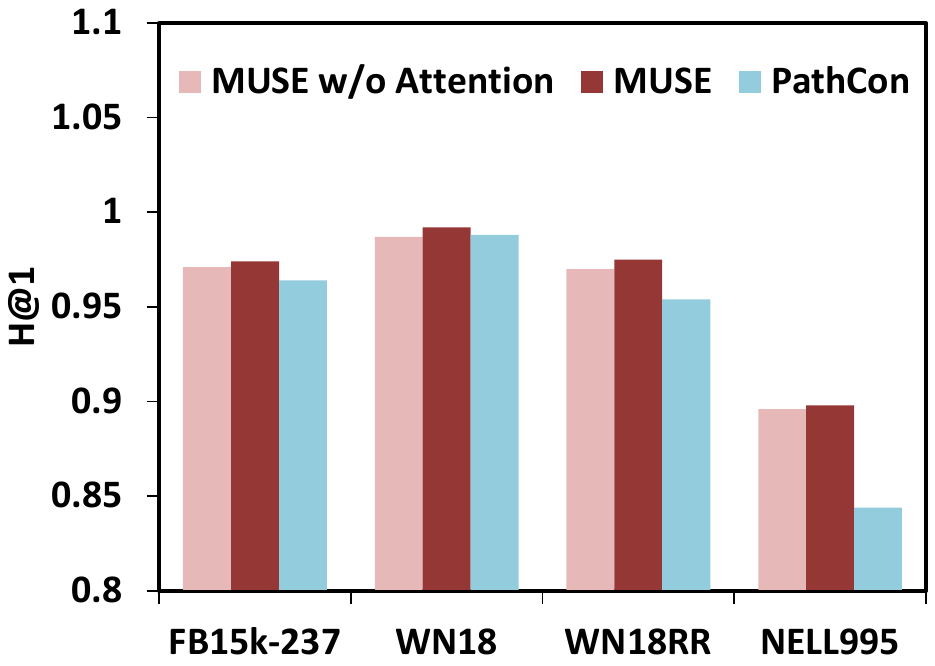}}
    \caption{Analysis of the Entity Description in Relation Prediction Task. 
    We define the $\text{\textbf{{MUSE~w/o~Fine-Tuning}}}$ model as applying the BERT model directly in Figure \ref{fig:pkla}. 
   As shown in Figure \ref{fig:pklb}, the $\text{\textbf{{MUSE w/o Attention}}}$ model aggregates entities without using attention mechanism in Equation \ref{eq:attention}.}
    \label{fig:pkl}
\end{figure*} 
\subsection{Contributions of Semantic Knowledge in Entity Representations}
In this subsection, we examine the role of semantic knowledge in entity representation during relation prediction.
As shown in Figure \ref{fig:pkl}, we analyze the impact of fine-tuning BERT in the prior knowledge learning, and edge-based attention mechanism in the contextual message passing.

Figure \ref{fig:pkla} shows that MUSE consistently exceeds the performance of PathCon on all datasets after vanilla BERT initializes the knowledge graph. Additionally, fine-tuning BERT significantly improves entity representations effectively and accurately, with a 0.034 H@1 increase on the NELL995 dataset.

Another analysis presented in Figure~\ref{fig:pklb} concentrates on the context message passing.
Similarly, the experiments show that MUSE on each dataset is higher than the model without the edge-based attention. It proves the effectiveness of semantics interaction in relation prediction. 
Besides, compared to these two semantic knowledge injection strategies, fine-tuning the language model can guide MUSE to learn sufficient entity representation and improve performance.
\begin{table*}[!t]
\small
\centering
\caption{Relation Prediction of MUSE and PathCon in LIS and RIS Scenarios.}
\resizebox{0.65\linewidth}{!}{
\begin{tabular}{l|l|c|c|c|c} 
\hline
\multirow{2}{*}{Test Scenarios}               & \multirow{2}{*}{Methods}       & FB15k-237              & WN18                   & WN18RR                                               & NELL995                 \\ 
\cline{3-6}
                                         &                                & H@1                    & H@1                    & H@1                                                  & H@1                     \\ 
\hline
\multirow{3}{*}{LIS Scenario} & PathCon                        & 0.917                  & 0.983                  & 0.878                                                & 0.719                   \\ 
\cline{2-6}
                                         & \multirow{2}{*}{\textbf{MUSE}} & \textbf{0.975~}        & \textbf{0.992}         & \textbf{0.948}                                       & \textbf{0.858}          \\
                                         &                                & ± 0.001                & ± 0.001                & ± 0.005                                              & ± 0.008                 \\ 
\hline
\multirow{3}{*}{RIS Scenario}    & PathCon                        & 0.962                  & 0.983                  & 0.967                                                & 0.897                   \\ 
\cline{2-6}
                                         & \multirow{2}{*}{\textbf{MUSE}} & \textbf{0.977}\uline{} & \textbf{0.991}\uline{} & \textcolor[rgb]{0.2,0.2,0.2}{\textbf{0.978}}\uline{} & \textbf{0.922}\uline{}  \\
                                         &                                & ± 0.001                & ± 0.001                & ± 0.002                                              & ± 0.001                 \\
\hline
\end{tabular}
}
\label{tab:lt}
\end{table*}

\begin{figure}[!t]
    \centering 
    \subfigure[Relation Prediction under Different Number of Triplet Paths.] { 
    \label{fig:4a} 
     \includegraphics[width=0.42\linewidth]{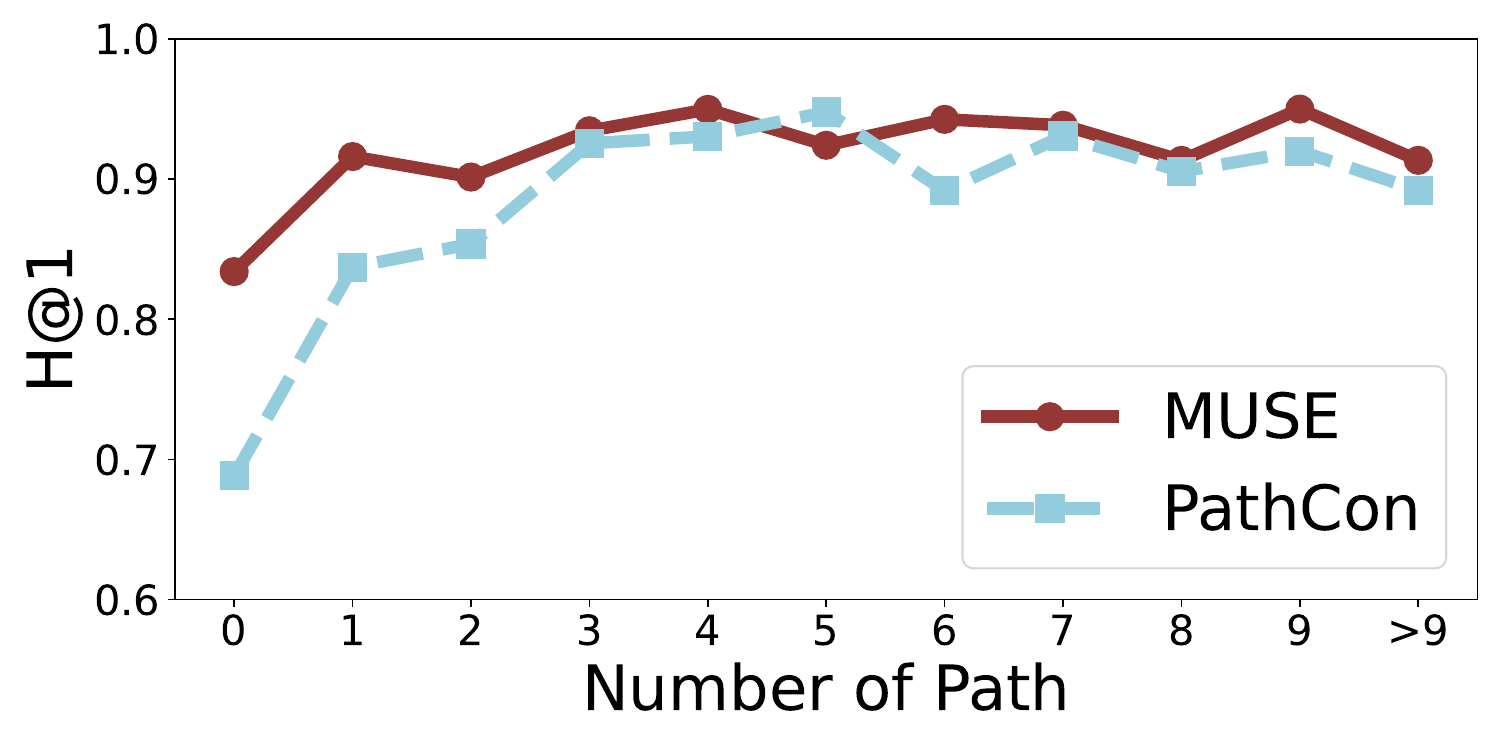}}
     \subfigure[Relation Prediction under Different Number of Entity Degrees.] { 
     \label{fig:4b}
    \includegraphics[width=0.42\linewidth]{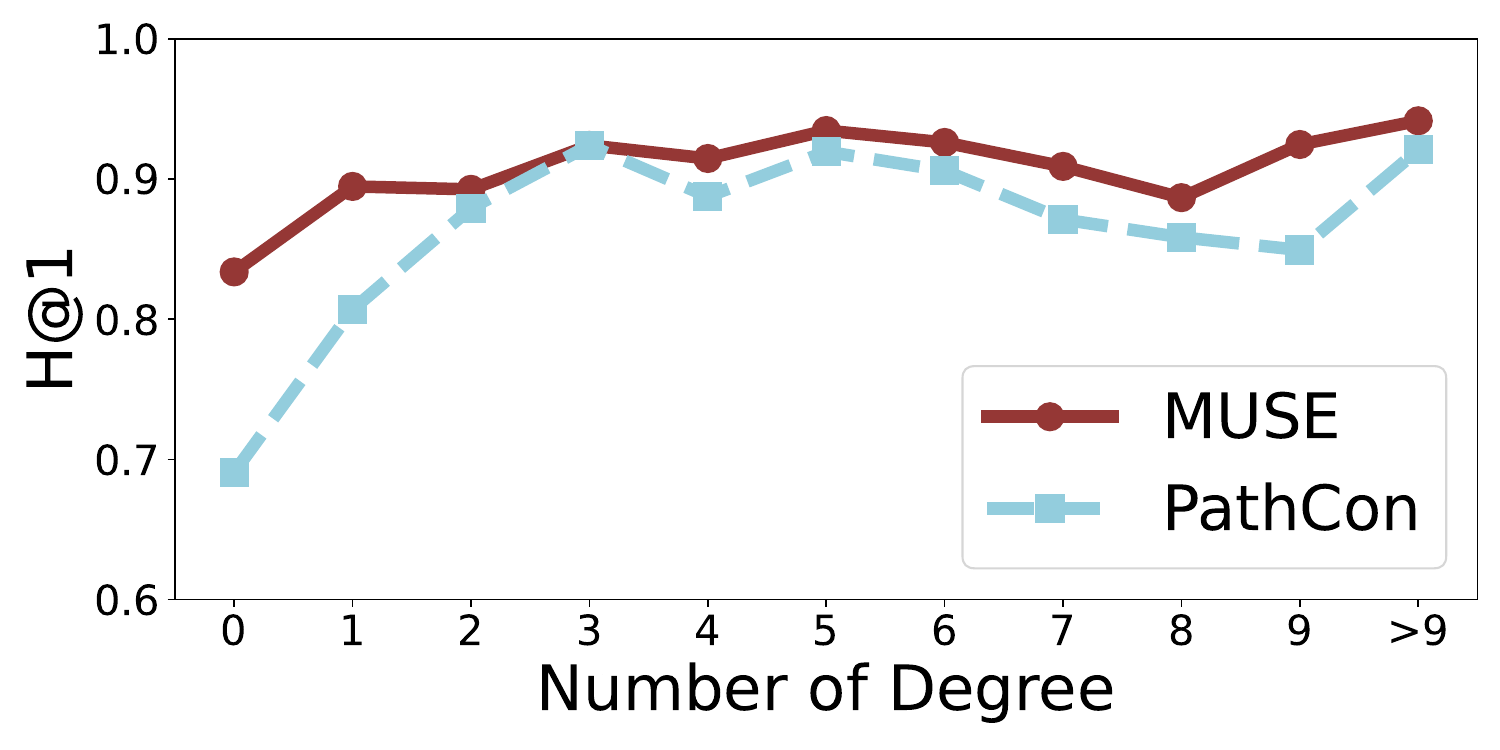}}
    \caption{Comparison of MUSE and PathCon Performance on the NELL995.}
    \label{fig:line}
\end{figure}
\subsection{Effectiveness of MUSE in the LIS and RIS Scenarios}
In this subsection, we investigate the effectiveness of our model and PathCon on all datasets with limited or rich graph information. 

MUSE outperforms the baseline in the relation prediction task in both LIS and RIS scenarios.
As shown in Table \ref{tab:lt}, MUSE demonstrates more significant advantages over PathCon in the LIS scenario. Notably, it achieves an improvement of 0.07 H@1 on the WN18RR dataset and 0.139 H@1 on the NELL995 dataset.
This increment is primarily due to the effective prior knowledge learning, which enhances the semantic representation and graph structure information.
External semantic knowledge is essential to enrich entity representation and improves relation prediction performance, especially for many long-tail entities.
Our experiments further analysis the impact of the prior knowledge learning on the NELL995 dataset. We compare the performance of MUSE and PathCon across various triplet paths and degrees in Figure~\ref{fig:4a} and Figure~\ref{fig:4b}, respectively. 
The results show that our model consistently outperforms PathCon in all experiments, especially when both paths and degrees are less than three.
\begin{figure}[t]
    \centering 
    \subfigure[Relationship Prediction from \textbf{\color{jeff}{Jeff Gordon}} (head entity) to \textbf{\color{race}{Auto Racing}} (tail entity) in the Knowledge Graph.] { 
    \label{fig:5a} 
     \includegraphics[width=0.85\linewidth]{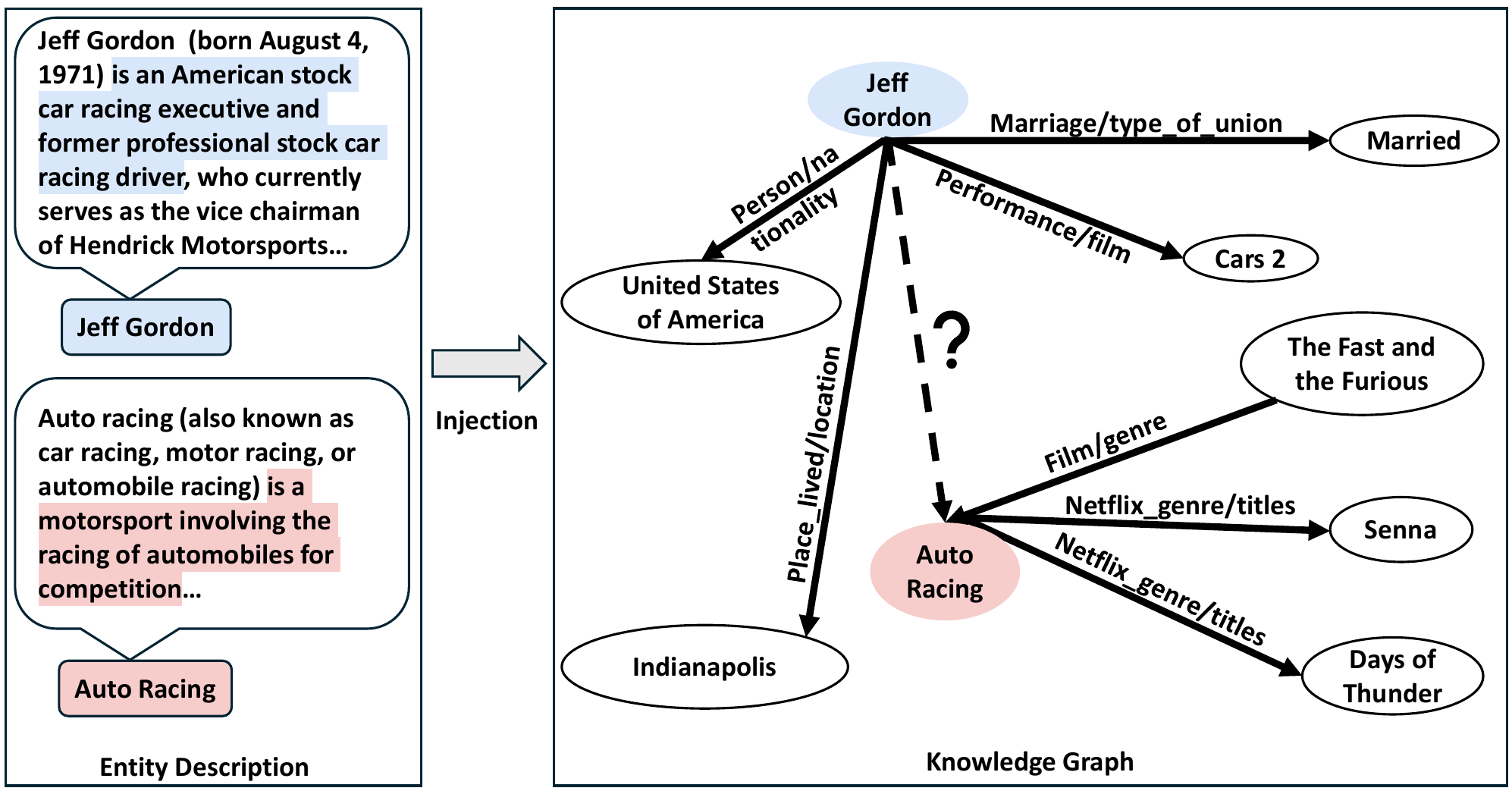}}
     \hfill
     \subfigure[Comparison of Relation Prediction Results of MUSE and PathCon.] { 
     \label{fig:5b}
    \includegraphics[width=0.85\linewidth]{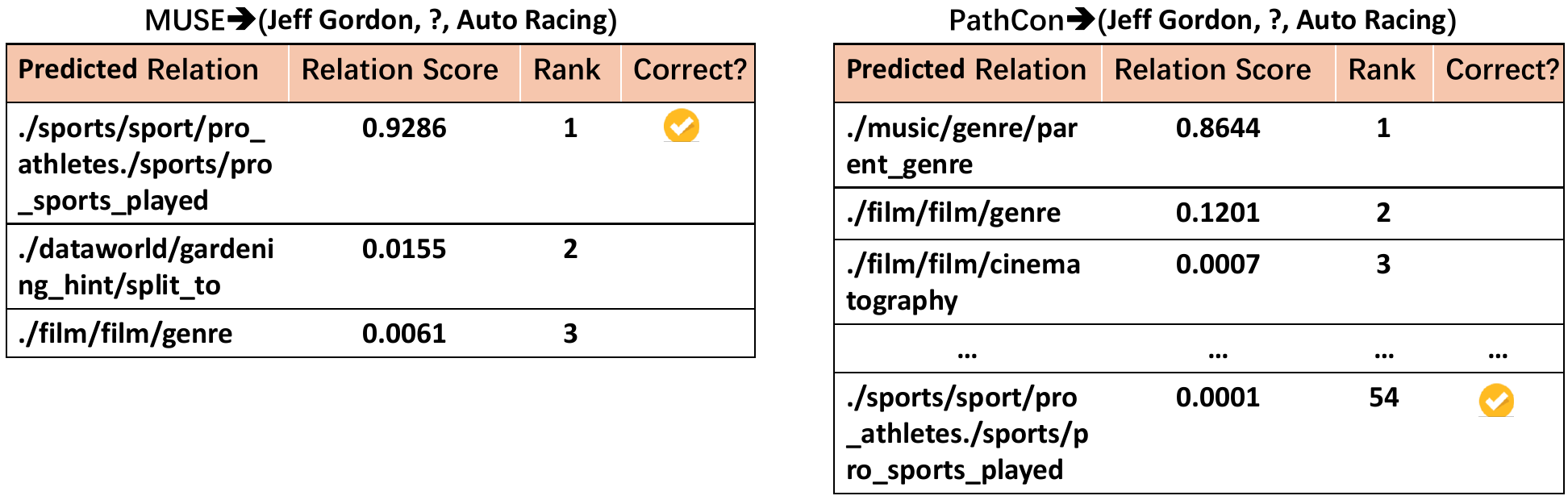}}
    \caption{An Easily Misjudged Example Case Predicted by MUSE and PathCon.}
    \label{fig:line}
\end{figure}
\subsection{Case Study}
This subsection demonstrates how the multi-knowledge integration works in an easily mispredicted relation prediction case.

In Figure~\ref{fig:5a}, we analyze a case study to predict the relation type from \textbf{\color{jeff}{Jeff Gordon}} (head entity) to \textbf{\color{race}{Auto Racing}} (tail entity). 
The connected entities and their paths suggest a film-related relation, making it hard to identify \textbf{\color{race}{Auto Racing}} as a sport. 
In this situation, PathCon mispredicts a high score for relations related to ``{./musid/genre/parent\_genre}'' and ranks the correct relation type at ${54_{th}}$ place. Meanwhile, MUSE correctly identifies the relation type and achieves the highest confidence score of 0.9286 in Figure \ref{fig:5b}. This is because the entity description states \textbf{\color{jeff}{Jeff Gordon}} as a race car driver and \textbf{\color{race}{Auto Racing}} as a car sport, injecting prior knowledge into the MUSE and helping to infer the  ``\textbf{./sports/sport/pro\_athletes./sports/pro\_sports\_played}'' relation. 

\section{Conclusion}
\label{sec:Conclusion}
We propose a novel model, MUSE, which Integrates three types of knowledge to enhance the entity representation: \textbf{Prior Knowledge Learning}, \textbf{Context Message Passing}, and \textbf{Relational Path Aggregation} for Knowledge Graph Completion.
Experiments indicate that MUSE outperforms conventional Knowledge Graph Completion models on four datasets.
The ablation study and the contribution evaluation of semantic knowledge have validated the importance of each knowledge. It is also shown that MUSE could perform well in both limited and rich information sets scenarios in the knowledge graph.
\bibliographystyle{splncs04}
\bibliography{refs}

\begin{thebibliography}{10}
\providecommand{\url}[1]{\texttt{#1}}
\providecommand{\urlprefix}{URL }
\providecommand{\doi}[1]{https://doi.org/#1}

\bibitem{transe}
Bordes, A., Usunier, N., Garcia-Duran, A., Weston, J., Yakhnenko, O.: Translating embeddings for modeling multi-relational data  (2013)

\bibitem{bordes2011learning}
Bordes, A., Weston, J., Collobert, R., Bengio, Y.: Learning structured embeddings of knowledge bases. In: Proceedings of AAAI. pp. 301--306 (2011)

\bibitem{COMET}
Bosselut, A., Rashkin, H., Sap, M., Choi, Y.: {COMET}: Commonsense transformers for automatic knowledge graph construction. In: Proceedings of ACL. pp. 4762--4779 (2019), \url{https://aclanthology.org/P19-1470}

\bibitem{cherniavskii-etal-2022-acceptability}
Cherniavskii, D., Artemova, E.: Acceptability judgements via examining the topology of attention maps. In: Findings of the Association for Computational Linguistics: EMNLP 2022 (2022), \url{https://aclanthology.org/2022.findings-emnlp.7}

\bibitem{Dettmers_Minervini_Stenetorp_Riedel_2018}
Dettmers, T., Minervini, P., Riedel, S.: Convolutional 2d knowledge graph embeddings  (2018), \url{https://ojs.aaai.org/index.php/AAAI/article/view/11573}

\bibitem{BERT-2019}
Devlin, J., Chang, M.W., Lee, K., Toutanova, K.: {BERT}: Pre-training of deep bidirectional transformers for language understanding. In: Proceedings of NAACL-HLT. pp. 4171--4186 (2019), \url{https://aclanthology.org/N19-1423}

\bibitem{conv}
Duvenaud, D., Maclaurin, D., Aguilera-Iparraguirre, J., Gómez-Bombarelli, R., Hirzel, T., Aspuru-Guzik, A., Adams, R.P.: Convolutional networks on graphs for learning molecular fingerprints (2015), \url{https://arxiv.org/pdf/1509.09292.pdf}

\bibitem{inductive}
Hamilton, W.L., Ying, R., Leskovec, J.: Inductive representation learning on large graphs (2018), \url{https://arxiv.org/pdf/1706.02216.pdf}

\bibitem{transd}
Ji, G., He, S., Xu, L., Liu, K., Zhao, J.: Knowledge graph embedding via dynamic mapping matrix. In: Proceedings of ACL-IJCNLP. pp. 687--696 (2015)

\bibitem{semi}
Kipf, T.N., Welling, M.: Semi-supervised classification with graph convolutional networks. arXiv preprint {\color{blue}\href{https://arxiv.org/abs/1609.02907}{arXiv:1609.02907}}  (2016)

\bibitem{Random}
Lao, N., Mitchell, T., Cohen, W.: Random walk inference and learning in a large scale knowledge base. In: Proceedings of EMNLP. pp. 529--539 (2011)

\bibitem{transr}
Lin, Y., Liu, Z., Sun, M.: Learning entity and relation embeddings for knowledge graph completion. In: Proceedings of AAAI (2015), \url{10.1609/aaai.v29i1.9491}

\bibitem{liu2024semdrsemanticawaredualencoder}
Liu, P., Zhang, W., Ding, Y., Zhang, X., Yang, S.H.: Semdr: A semantic-aware dual encoder model for legal judgment prediction with legal clue tracing (2024)

\bibitem{nayyeri2021trans4e}
Nayyeri, M., Motta, E., et~al.: Trans4e: Link prediction on scholarly knowledge graphs. Neurocomputing  \textbf{461},  530--542 (2021)

\bibitem{node2vec}
Palumbo, E., Rizzo, G., Troncy, R., Baralis, E., Osella, M., Ferro, E.: Knowledge graph embeddings with node2vec for item recommendation. In: Proceedings of ESWC (2018), \url{https://doi.org/10.1007/978-3-319-98192-5_22}

\bibitem{drum}
Sadeghian, D.Z.: Drum: End-to-end differentiable rule mining on knowledge graphs  (2019), \url{https://dl.acm.org/doi/abs/10.5555/3454287.3455662}

\bibitem{text2}
Sarzynska-Wawer, J., Okruszek, L.: Detecting formal thought disorder by deep contextualized word representations. Psychiatry Research  (2021)

\bibitem{kgc_review_1}
Shen, T., Zhang, F., Cheng, J.: A comprehensive overview of knowledge graph completion. Knowledge-Based Systems p. 109597 (2022), \url{https://www.sciencedirect.com/science/article/abs/pii/S095070512200805X}

\bibitem{Towards}
Shomer, H., Jin, W., Wang, W., Tang, J.: Toward degree bias in embedding-based knowledge graph completion. In: Proceedings of WWW. p. 705–715. New York, NY, USA (2023), \url{https://doi.org/10.1145/3543507.3583544}

\bibitem{rotate}
Sun, Z., Deng, Z.H., Nie, J.Y., Tang, J.: Rotate: Knowledge graph embedding by relational rotation in complex space. arXiv preprint {\color{blue}\href{https://arxiv.org/abs/1902.10197}{arXiv:1902.10197}}  (2019)

\bibitem{observed}
Toutanova, K., Chen, D.: Observed versus latent features for knowledge base and text inference. In: Proceedings of the Workshop on Continuous Vector Space Models and their Compositionality (2015), \url{https://aclanthology.org/W15-4007}

\bibitem{fusion1}
Toutanova, K., Chen, D., Pantel, P., Poon, H., Choudhury, P., Gamon, M.: Representing text for joint embedding of text and knowledge bases. In: Proceedings of EMNLP. pp. 1499--1509 (2015), \url{https://aclanthology.org/D15-1174.pdf}

\bibitem{distmult}
Toutanova, K., Gamon, M.: Representing text for joint embedding of text and knowledge bases. In: Proceedings of EMNLP. pp. 1499--1509 (2015)

\bibitem{complex}
Trouillon, T., Welbl, J., Riedel, S., Gaussier, {\'E}., Bouchard, G.: Complex embeddings for simple link prediction. In: Proceedings of ICML. pp. 2071--2080 (2016)

\bibitem{doloers}
Wang, H., Kulkarni, V., Wang, W.Y.: Dolores: Deep contextualized knowledge graph embeddings  (2018), \url{https://arxiv.org/pdf/1811.00147.pdf}

\bibitem{pathcon}
Wang, H., Ren, H., Leskovec, J.: Relational message passing for knowledge graph completion. In: Proceedings of KDD. pp. 1697--1707 (2021)

\bibitem{kgc_review_2}
Wang, M., Qiu, L., Wang, X.: A survey on knowledge graph embeddings for link prediction. Symmetry p.~485 (2021), \url{https://www.mdpi.com/2073-8994/13/3/485}

\bibitem{transh}
Wang, Z., Zhang, J., Feng, J., Chen, Z.: Knowledge graph embedding by translating on hyperplanes. In: Proceedings of AAAI (2014), \url{10.1609/aaai.v28i1.8870}

\bibitem{KICGPT}
Wei, Y., Huang, Q., Zhang, Y., Kwok, J.: {KICGPT}: Large language model with knowledge in context for knowledge graph completion. In: Findings of EMNLP. pp. 8667--8683 (2023), \url{https://aclanthology.org/2023.findings-emnlp.580}

\bibitem{fusion2}
Xie, R.: Representation learning of knowledge graphs with entity descriptions. In: Proceedings of AAAI (2016). \doi{10.1609/aaai.v30i1.10329}

\bibitem{nell}
Xiong, W., Hoang, T., Wang, W.Y.: Deeppath: A reinforcement learning method for knowledge graph reasoning. arXiv preprint {\color{blue}\href{https://arxiv.org/abs/1707.06690}{arXiv:1707.06690}}  (2017)

\bibitem{graphvite}
Yang, B., Yih, W.t., He, X., Gao, J., Deng, L.: Embedding entities and relations for learning and inference in knowledge bases. arXiv preprint {\color{blue}\href{https://arxiv.org/abs/1412.6575}{arXiv:1412.6575}}  (2014)

\bibitem{neurip}
Yang, F., Yang, Z., Cohen, W.W.: Differentiable learning of logical rules for knowledge base reasoning \url{https://dl.acm.org/doi/abs/10.5555/3294771.3294992}

\bibitem{kgbert}
Yao, L., Mao, C., Luo, Y.: Kg-bert: Bert for knowledge graph completion. arXiv preprint {\color{blue}\href{https://arxiv.org/abs/1909.03193}{arXiv:1909.03193}}  (2019)

\bibitem{pre_1}
Yasunaga, M., Liang, P.S., Leskovec, J.: Deep bidirectional language-knowledge graph pretraining. In: Proceedings of NeurIPS (2022)

\bibitem{ali}
Zeng, K., Li, C., Hou, L., Li, J., Feng, L.: A comprehensive survey of entity alignment for knowledge graphs. AI Open  \textbf{2},  1--13 (2021)

\bibitem{longtail_v1}
Zhang, N., Deng, S., Sun, Z., Wang, G., Chen, X., Zhang, W., Chen, H.: Long-tail relation extraction via knowledge graph embeddings and graph convolution networks. arXiv preprint  (2019), \url{https://arxiv.org/pdf/1903.01306.pdf}

\bibitem{quate}
Zhang, S., Tay, Y., Yao, L., Liu, Q.: Quaternion knowledge graph embeddings  \textbf{32} (2019), \url{https://dl.acm.org/doi/abs/10.5555/3454287.3454533}

\bibitem{zhang2020relational}
Zhang, Z., He, Q.: Relational graph neural network with hierarchical attention for knowledge graph completion. In: Proceedings of AAAI (2020)

\bibitem{ERNIE}
Zhang, Z., Liu, Q.: {ERNIE}: Enhanced language representation with informative entities. In: Proceedings of ACL (2019), \url{https://aclanthology.org/P19-1139}

\end{thebibliography}
\end{document}